\documentclass{article}
\usepackage{float}
\usepackage{subcaption}
\usepackage{graphicx}
\usepackage{colortbl}
\usepackage{comment}

\usepackage[preprint,nonatbib]{arXiv_VideoSummarization}
\usepackage{amsmath}

\usepackage{ulem} 

\usepackage[utf8]{inputenc} 
\usepackage[T1]{fontenc}    
\usepackage{hyperref}       
\usepackage{url}            
\usepackage{booktabs}       
\usepackage{amsfonts}       
\usepackage{nicefrac}       
\usepackage{microtype}      
\usepackage{xcolor}         
\usepackage{booktabs}
\usepackage{multirow}
\usepackage{pifont}

\hypersetup{
    colorlinks=true,
    citecolor=blue,   
    linkcolor=blue,   
    urlcolor=blue     
}

\newcommand{\xmark}{\ding{55}}

\title{TRIM: A Self-Supervised Video Summarization Framework Maximizing Temporal Relative Information and Representativeness}

\author{%
  Pritam Mishra$^{1}$ \\
  Pompeu Fabra University\\
  \texttt{pritam.mishra@upf.edu}
  \And
  Coloma Ballester$^{1}$\\
  Pompeu Fabra University\\
  \texttt{coloma.ballester@upf.edu}
  \And
  Dimosthenis Karatzas$^{2}$\\
  Universitat Autònoma de Barcelona\\
  \texttt{dimos@cvc.uab.es}}

\begin{document}

\maketitle

\begin{abstract}
The increasing ubiquity of video content and the corresponding demand for efficient access to meaningful information have elevated video summarization and video highlights as a vital research area. However, many state-of-the-art methods depend heavily either on supervised annotations or on attention-based models, which are computationally expensive and brittle in the face of distribution shifts that hinder cross-domain applicability across datasets. We introduce a pioneering self-supervised video summarization model that captures both spatial and temporal dependencies without the overhead of attention, RNNs, or transformers. Our framework integrates a novel set of Markov process-driven loss metrics and a two-stage self supervised learning paradigm that ensures both performance and efficiency. Our approach achieves state-of-the-art performance on the SUMME and TVSUM datasets, outperforming all existing unsupervised methods. It also rivals the best supervised models, demonstrating the potential for efficient, annotation-free architectures. This paves the way for more generalizable video summarization techniques and challenges the prevailing reliance on complex architectures.
\end{abstract}

\footnotetext{Code and pretrained models are available at: \url{https://github.com/pmishra-ai/TRIM}}
\section{Introduction}
The explosive growth of visual media, particularly long-form video content, has created an urgent need for methods that can efficiently organize, retrieve, and understand large-scale video data. From online streaming platforms and educational archives to surveillance systems and personal media collections, users are increasingly confronted with volumes of video that far exceed the time available for full consumption. As attention spans shorten and information density increases, the ability to transform raw videos into concise yet semantically meaningful summaries has become a central challenge in computer vision and multimedia analysis. Video summarization addresses this problem by selecting the most informative and representative segments of a video, allowing users to access essential content quickly without exhaustive viewing.

Despite substantial progress, many state-of-the-art video summarization approaches remain heavily dependent on supervised annotations, such as frame-level importance scores or user-generated summaries. While effective, these methods are inherently constrained by the availability, cost, and subjectivity of labeled data. Human annotations are expensive to obtain at scale, often inconsistent across annotators, and may not generalize well to new domains, video styles, or unseen content distributions. These limitations motivate the search for learning paradigms that can infer meaningful summaries directly from the intrinsic structure of video data, without relying on manual supervision.

A fundamental difficulty in video summarization lies in modeling long-range temporal dependencies while simultaneously preserving local visual context. Important events may span distant portions of a video, and semantically relevant frames are not always temporally adjacent. Existing methods often employ recurrent networks, self-attention, graph neural networks, or hybrid architectures to capture such dependencies. Although powerful, these approaches can introduce substantial computational overhead and often require complex training procedures. Moreover, methods that focus primarily on temporal relations may overlook the rich spatial structure contained within individual frames.

In this paper, we propose a different perspective: instead of relying on supervision or expensive global attention mechanisms, we formulate video summarization through \emph{Markov process-driven self-supervision}. We introduce a new family of information-theoretic and transition-based metrics that act as unsupervised training signals for identifying informative and representative video frames. These objectives are designed to capture temporal coherence, dependency structure, and summary quality directly from sequential visual data.

Building on this principle, we develop a simple yet highly effective architecture based on lightweight 1D convolutional networks that model spatio-temporal relationships across frame sequences. Unlike many recent approaches, the proposed method does not require transformers, recurrent networks, or supervised annotations. Nevertheless, it captures both temporal evolution and local contextual structure efficiently, making it well suited for large-scale or resource-constrained settings.

Empirically, the proposed framework achieves state-of-the-art performance on standard benchmarks including SUMME~\cite{summe_dataset} and TVSUM~\cite{tvsum_dataset}. Remarkably, strong results are obtained not only in the full two-stage self-supervised setting, but also when training solely with the second-stage objective, outperforming prior unsupervised approaches.

The main contributions of this chapter are summarized as follows:

\begin{itemize}
    \item We propose a simple and computationally efficient framework for video summarization that leverages 1D CNNs to capture spatio-temporal dependencies in a fully self-supervised manner.
    
    \item We introduce a two-stage self-supervised learning strategy, while also showing that the second stage alone is sufficient to achieve highly competitive and state-of-the-art unsupervised performance.
    
    \item We develop novel Markov process-based metrics and representativeness objectives that serve as principled unsupervised loss functions for predicting frame importance.
    
    \item We demonstrate state-of-the-art results on the SUMME~\cite{summe_dataset} and TVSUM~\cite{tvsum_dataset} benchmarks, validating the effectiveness of simple architectures guided by strong self-supervised objectives.
\end{itemize}

\section{Related Work}\label{sec:relwork}
Video summarization methods can be broadly categorized into supervised, unsupervised and self-supervised approaches. Unsupervised video summarization has leveraged a range of methods to generate concise and informative summaries without requiring labeled data. Reinforcement learning-based methods, such as the approach proposed in \cite{zhou2018deep}, aim to maximize diversity and representativeness of summarized videos through novel design of reward functions. Traditional machine learning algorithms, including clustering and sparse dictionary learning, have also been extensively utilized, with $L_{2,0}$-constrained sparse dictionary learning applied in \cite{sparse_dict_2014} as a notable example. More recently, adversarial models like SGAN \cite{gansum} have been introduced, using generative networks to produce summaries that are nearly indistinguishable from the original videos.

Most supervised video summarization methods rely on deep neural networks to capture temporal dependencies \cite{stvt,dsnet,a2summ,vasnet,maam,rrstg,msva,hsa,vjmht,dmasum} and typically require annotated human-generated summaries for training. Recent methods increasingly leverage attention mechanisms to enhance the efficiency and quality of video summarization. Models like VASNet~\cite{vasnet}, SUM-GDA~\cite{sum-gda}, and CA-SUM~\cite{CA-SUM} progressively improve attention-based summarization by incorporating efficiency, diversity, and uniqueness, respectively. Other models such as DSNet \cite{dsnet} and PGL-SUM~\cite{PGLSUM} use attention to refine shot localization and summary precision by leveraging transformer networks, VJMHT \cite{vjmht} improves performance through the learning of semantic similarities across related videos.

Self-supervised learning \cite{barlow,byol,simclr} has been widely used to pre-train deep models to transfer the representation learned towards several downstream tasks. Based on the recent success of self-supervised learning, it has also been exploited for video summarization domain. Recently, CLIP-It \cite{clipit} achieves high performance by combining CLIP-extracted frame features with a six-layer Transformer, though this approach incurs substantial computational cost during inference \cite{sspvs}. Recently, SSPVS \cite{sspvs} has exploited multi-modal self-supervised learning on a large video dataset to generate a progressive video summarization by stacking multiple models with identical architecture.

\section{Proposed Method}\label{sec:method}
In this section, we introduce a novel two-stage self-supervised learning method for summarizing videos. To that goal, we first propose in Sect.~\ref{ssec:2metrics} a new class of Markov process-driven metrics, that are then used in Sects.~\ref{ssec:2stagesmethod}, \ref{sec:ssl-pre-training} and \ref{ssect:stage2} as unsupervised loss signals, that capture semantic information progressively within video content without reliance on manual annotations. The first stage of our method is defined in Sect.~\ref{sec:ssl-pre-training} and the second stage in Sect.~\ref{ssect:stage2}. 

First, let us define the notation. For each video \( V \), let \( N \) be the number of image frames. Each image frame will be denoted by \( I_t \), where $t\in\{1,\dots,N\}$. Using a frozen deep CNN \( f \), we extract feature representations for the video frames. Namely, 
$X_t = f(I_t)$ represents the feature vector of the corresponding image frame at time step \( t \). Our method  learns an importance score $p_t$ for each frame $I_t$. From the importance scores $\{p_t\}_{t=1}^N$ we obtain the video summary.


\subsection{Proposed underlying metrics}\label{ssec:2metrics}
In our proposed metrics, instead of comparing image features within a video \( V \) based on their temporal order, we take a different approach by analyzing the changes in feature distributions over sequential time steps. The feature distribution at the \( t \)-th time step is computed using the softmax function applied across the feature dimensions, namely, 
$D_t^j = \frac{e^{X_t^j}}{\sum_{l=1}^{d} e^{X_t^l}}$.
where \( X_t \in \mathbb{R}^{d} \) is the feature vector of  $I_t$ at time  \( t \), \( d \) is the number of feature dimensions (e.g., 2048), and \( X_t^j\in \mathbb{R} \) represents the \( j \)-th feature of \( X_t \). This formulation aims to transform the feature values into a probability distribution, allowing for a meaningful comparison of changes in feature distributions over time.

\subsubsection{Pairwise Temporal Relative Information}
We propose a novel metric based on the entropy, or the amount of information, contained within these distributions over time. Our metric quantifies the pairwise temporal change in information between consecutive time steps and is defined as
\begin{equation}
    \Delta_t = \left| \frac{H_t - H_{t-1}}{H_t} \right|
    \label{eq:ptri}
\end{equation}
where the entropy $H_t$ of the feature probability distribution \( D_t \) at time  \( t \) is defined as always as:  
$H_t = H(D_t) = - \sum_{j=1}^{d} D_t^j \log D_t^j$,
where \( D_t^j \) is the \( j \)-th element of  \( D_t \). 
In the following, we refer to $\Delta_t$ in \eqref{eq:ptri} as PTRI metric, where PTRI stands for Pairwise Temporal Relative Information.

\subsubsection{Pairwise Contextual Temporal Relative Information}
While our first metric PTRI ($\Delta_t$ in \eqref{eq:ptri}) quantifies the relative amount of information change at the current time step compared to the previous one, we introduce also a metric that measures the total information change from the current time step to all preceding time steps. In particular, we define the Pairwise Contextual Temporal Relative Information (PCTRI) metric as
\begin{equation}
    \Gamma_t = \left| \frac{H_t - \frac{1}{t-1}\sum\limits_{i=1}^{t-1}{H_{i}}}{H_t} \right|
    \label{eq:pctri}
\end{equation}
This measure captures the contextual change in information relative to the current state.
Unlike previous works such as \cite{zhou2018deep} that compare a state to all others without accounting for sequential dependencies, our metrics, PTRI and PCTRI, explicitly capture both Markovian and history-dependent structures, respectively. 

\subsection{Proposed Two-Stages Strategy for a Self-Supervised Learning Paradigm}\label{ssec:2stagesmethod}



We briefly introduce our data pipeline for the proposed method. Given a video \( V \) made of image frames \( \{ I_1, I_2, \dots, I_N \} \), as mentioned before we extract frame features using a frozen pre-trained CNN \( f \) (ResNet-50 in this work), resulting in feature representations \( \{ X_1, X_2, \dots, X_N \} \), where $X_t = f(I_t)$. The proposed training pipeline can be delineated into two distinct stages.

\begin{figure}[htbp]
  \centering
  \includegraphics[width=\textwidth]{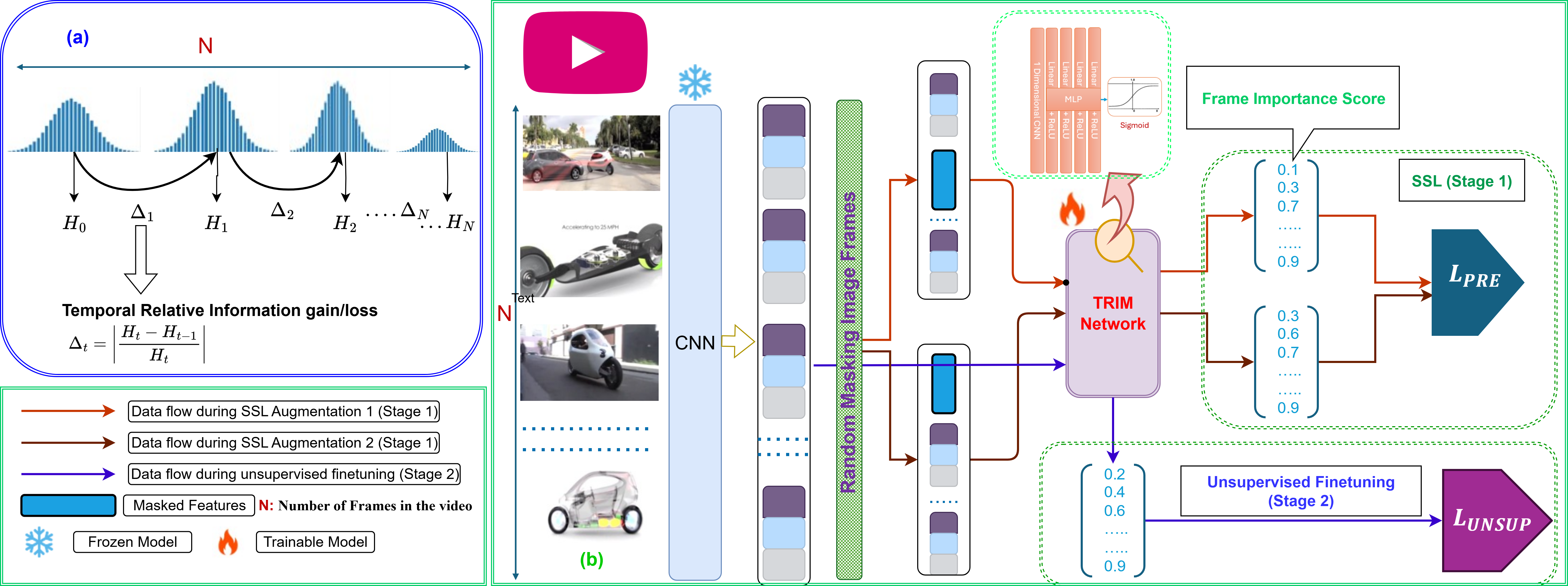}
    \caption{
\textcolor{blue}{\textbf{(a)}} Overview of the PTRI metric. 
\textcolor{green}{\textbf{(b)}} Pipeline of the proposed two-stage self-supervised framework. 
Red and brown arrows indicate the Stage~1 self-supervised pre-training process, while blue arrows represent the Stage~2 unsupervised fine-tuning stage.
}
  \label{fig:ssl_overview}
\end{figure}

\paragraph{Stage 1: Self-Supervised Pre-training:} We propose a self-supervised pre-training framework to learn noise invariant robust visual representations from videos. To that goal, we propose to randomly mask m\% of the features \( \{ X_1, X_2, \dots, X_N \} \) and create two augmented views to learn robust video representations through our proposed network. The training pipeline for SSL pre-training is outlined in Fig.~\ref{fig:ssl_overview} and is described in Sect.~\ref{sec:ssl-pre-training}.

\paragraph{Stage 2: Unsupervised fine-tuning:} Stage 2 leverages the weights from pre-training to initialize the network, which is subsequently fine-tuned using the unsupervised loss functions described in Section~\ref{ssect:stage2}. All extracted features for a given video are processed in a single batch through our proposed neural network described in Fig.~\ref{fig:ssl_overview}, which generates an {importance score} or {selection probability} \( p_t \) for each frame, where \( t \) represents the time step within the video (\( 1 \leq t \leq N \)). These scores are used to compute our {temporal relative information losses} and {representative loss} without requiring discrete frame selection unlike in a reinforcement learning framework.  

It is important to note that our proposed method can operate in an entirely unsupervised manner using only Stage 2, without requiring pre-training. A performance comparison between Stage 2 alone and the full two-stage approach is presented in Table~\ref{tab:ablation}, with a detailed discussion of each component provided in the following sections.

\paragraph{Network Architecture} We propose a simple yet highly efficient neural network architecture to demonstrate the efficacy of our methodology. The network consists of a single 1D convolutional layer followed by linear layers with ReLU activation functions, followed by a final sigmoid layer that maps the output to a probability $p_t$ at each time step \( t \). Compared to existing state-of-the-art methods, our network can effectively capture spatio-temporal relationship without  complex architectures such as RNN \cite{dpplstm,zhou2018deep}, GAN\cite{gansum,gansum_aae}, Attention\cite{csta,sspvs} mechanism or Transformers \cite{clipit,ma}.

More precisely, our architecture comprises a  single 1D convolutional layer of kernel size 3, padding size of 1 followed by a series of fully connected layers: \textbf{\(  \mathbf{Conv1D[2048} \to \mathbf{1024]} \to \mathbf{FC[1024} \to \mathbf{512]} \to \mathbf{FC[512} \to \mathbf{256]} \to \mathbf{FC[256} \to \mathbf{1]} \)}, with ReLU activations after each layer except the final one, which applies a sigmoid function.

\subsection{Stage 1: Self-Supervised Pre-training}
\label{sec:ssl-pre-training}
Inspired by \cite{ma,sspvs}, we introduce a  Self-Supervised Learning (SSL) framework as the first pre-training stage that, unlike \cite{ma,sspvs}, our approach does not rely on an autoencoder or transformer architecture. In fact, most contrastive and SSL frameworks do not require an encoder-decoder architecture, as seen in \cite{barlow,simclr,byol}. This suggests that a simple encoder with a well designed loss is sufficient to learn robust representations. Thus, We introduce a loss function incorporating a Pearson correlation-based term that encourages consistency and robustness by maximizing agreement between two noise-augmented views of the same video representation. Additionally, we increase the standard deviation of the importance scores to amplify the differences between augmented views.




More precisely, we propose a data augmentation strategy where we randomly mask \(m\%\) of frames from the feature set \(X\) within a video \(V\). The resulting masked feature set is denoted as \(X'\) (that we refer in Fig.~\ref{fig:ssl_overview} to as Augmentation View 1), where the selected masked frames are replaced with zeros. We apply the same masking process a second time to obtain \(X''\) (Augmentation View 2 in Fig.~\ref{fig:ssl_overview}). 
The masking ratio \(m\) is randomly chosen between 15\% and 50\%, and a performance comparison for different values of \(m\) is provided in the {ablation studies}(Sect.~\ref{ssec:ablations}, (Fig.~\ref{fig:nu_mask_ratio}). 

Next, we pass \(X'\) and \(X''\) through our  network to obtain \( \mathbf{p}_t' = [p_1', p_2', \dots, p_n'] \) and \( \mathbf{p}_t'' = [p_1'', p_2'', \dots, p_n''] \), representing probabilities of selecting the \(t\)-th frame in the video, for all \( t \). 

Finally, our proposed loss function during SSL(stage 1) is defined as:
\begin{equation}\label{eq:lossPre}
    L_\text{PRE} = L_\text{CORR} + \nu (L_\text{SD1} + L_\text{SD2}) ,
\end{equation} 
\begin{equation}
L_\text{CORR} = 1 - \frac{\sum_{t=1}^{N} (p_t' - \bar{p}')(p_t'' - \bar{p}'')}{\sqrt{\sum_{t=1}^{N} (p_t' - \bar{p}')^2} \cdot \sqrt{\sum_{t=1}^{N} (p_t'' - \bar{p}'')^2}}
\end{equation}
where $\bar{p}' = \frac{1}{N} \sum_{t=1}^{N} p_t'$  is the mean of vector \( \mathbf{p}_t' \),  $\bar{p}'' = \frac{1}{N} \sum_{t=1}^{N} p_t''$ is the mean of vector \( \mathbf{p}_t'' \). 
Finally,
\begin{equation}
        L_\text{SD1} = \frac{1}{\sqrt{\frac{1}{N}\sum\limits_{t=1}^{N}(p'_t - \bar{p'})^2}} 
\quad \text{and} \quad
        L_\text{SD2} = \frac{1}{\sqrt{\frac{1}{N}\sum\limits_{t=1}^{N}(p''_t - \bar{p''})^2}}
\end{equation}
In \eqref{eq:lossPre}, $\nu$ is an hyper-parameter set to $0.005$ and $0.0005$, respectively, for TVSUM and SUMME datasets in our experimental results in Sect.~\ref{sec:experiments}. For an ablation study we refer the readers to Fig.~\ref{fig:hyperparam_nu}.

\subsection{Stage 2: Unsupervised fine-tuning}\label{ssect:stage2}
In this stage, we process all features from the video without any masking or augmented views through our proposed network by transferring the weights from the pre-training stage and fine-tuning with our unified unsupervised loss function $L_\text{UNSUP}$, defined as follows.\\
\textbf{Pairwise Temporal Relative Information Maximization Loss: }
The underlying principle of this loss  is to maximize the pairwise relative information between consecutive distributions, thereby assigning higher probabilities to scenes that differ significantly from their predecessors. This loss ensures that frames which do not show significant changes are assigned lower probabilities. We derive this loss from equation \eqref{eq:ptri} as follows,
\begin{equation}
    L_\text{PTRIM} = \frac{N-1}{\sum\limits_{t=2}^{N} \, p_t \,\Delta_t}
\end{equation}

\paragraph{Pairwise Contextual Temporal Relative Information Maximization Loss: }
The goal of this loss function is to enhance the information diversity within a video by maximizing the relative difference between each frame’s distribution and all preceding distributions. This approach ensures that the model prioritizes scenes with significant novel content, thereby increasing the "surprise" or new information at each time step. By encouraging greater variability between consecutive frames, the method emphasizes the importance of capturing significant transitions in the video, rather than repetitive or uninformative content. We can derive this loss function using equation \eqref{eq:pctri} as follows,
\begin{equation}
    L_\text{PCTRIM} = \frac{N-1}{\sum\limits_{t=2}^{N}\, p_t \, \Gamma_t}
\end{equation}

\paragraph{Representation Loss: }
Building on the approach proposed in \cite{zhou2018deep}, we introduce a representative loss alongside information diversity losses, as previously discussed, for video summarization. While most state-of-the-art methods employ cosine similarity between image frames as a representative reward within a reinforcement learning framework, we argue that cosine similarity fails to capture the nonlinear and complex relationships between original frames and the predicted summarized frames. To address this limitation, we propose leveraging the Wasserstein-2 distance \cite{villani2008optimal} between original video frames against the predicted summarized frames by the network. 

Let $X=\{ X_1, X_2, \dots, X_N \}$ be the set of feature vectors and 
$Y = \{ p_1 X_1, p_2 X_2, \dots, p_N X_N \}$,
where \( p_t \) (for \( t = 1,\dots,N\)) denotes the probability of selecting each frame. Let us assume that $X_t$ and $Y_t=p_t Xt$ are, respectively, samples of two empirical distributions $X$ and $Y$, respectively. We define our representativeness loss $L_\text{REP}$ as the Wasserstein-2 distance \cite{villani2008optimal} between their probability distributions. In practice, we approximate the optimal transport problem using the Sinkhorn algorithm and, in particular, using the formulation of~\cite{geomloss}. The corresponding Wasserstein-2 distance is then given by
\begin{equation}
W_2^2(a, b) = \min_{\gamma \in \Pi(a, b)} \sum\limits_{i=1}^{N} \sum\limits_{j=1}^{N} \gamma_{ij} \| X_i - p_j X_j \|^2
\end{equation}
where 
    \( \gamma_{ij} \) represents the transport plan between frames, which is computed using Sinkhorn regularization~\cite{geomloss},
    and \( \Pi(a, b) \) is the set of valid couplings ensuring that the marginals align with the probability distributions \( a = \{ a_1, a_2, \dots, a_N \} \) and \( b = \{ p_1, p_2, \dots, p_N \} \).

The Sinkhorn algorithm introduces a regularization term to make the transport plan more computationally feasible and allows us to compute an approximate solution efficiently. We adopted the approach outlined in \cite{geomloss} for the efficient formulation of Sinkhorn divergence leveraging GPU acceleration.

\paragraph{Selection Diversity Loss: }
The sigmoid layer at the end of the proposed network outputs the probability of selecting a frame $p_t$ or importance score. However, if the pre-sigmoid linear layer produces sufficiently large values, the resulting probabilities can become close to 0.9 or 1. In order to avoid having similar importance scores, we introduce a diversity term that encourages the network to amplify distinction within importance scores. We ensured this by increasing the standard deviation of the importance score that distinguishes between consequential and trivial frames. The proposed selection diversity term is defined as


\begin{equation}
    L_\text{SD} = \frac{1}{\sqrt{\frac{1}{N}\sum\limits_{t=1}^{N}(p_t - \bar{p})^2}}
\end{equation}
where $p_t$ is the probability of selecting a frame at step $t$ and $\bar{p}$ is the mean probability of all $p_t$.

\paragraph{Leveraging Multiple Objectives in a Unified Loss Function: }
In conclusion, we present our unified unsupervised loss function, which simultaneously optimizes multiple objectives, as follows, 
\begin{equation}\label{eq:mainloss}
L_\text{UNSUP} =  \alpha L_\text{PTRIM} +  \beta L_\text{PCTRIM} +  \gamma L_\text{REP} + L_\text{SD}
\end{equation}
where $L_\text{REP}$ is defined as $W_2^2(X,Y)$. The values for $\alpha$, $\beta$, and $\gamma$ were set to 1, 5, and 0.5, respectively. For a detailed ablation study on hyperparameter selection and sensitivity, we refer to Sect.~\ref{ssec:ablations}.

\section{Experimental Results}\label{sec:experiments}
We compare TRIM with existing state-of-the-art (SOTA) methods on two widely adopted benchmark datasets: SUMME~\cite{summe_dataset} and TVSUM~\cite{tvsum_dataset}. The comparative results are presented in Table~\ref{tab:sota_comparison}. The results demonstrate that TRIM not only outperforms all other methods in the unsupervised and self-supervised categories, but also achieves performance comparable to the best supervised approaches. Remarkably, our method attains this level of performance with significantly lower computational cost (3.27 GFLOPs) and without relying on attention mechanisms or transformer-based architectures, unlike other SOTA methods. The computational efficiency of TRIM, relative to other approaches, is highlighted in the last column of Table~\ref{tab:sota_comparison}.
Additionally, we also report F1 score for our proposed method in \ref{ssec:additional_results}.
\begin{table}[h]
    \centering
    \renewcommand{\arraystretch}{1.2}
    \small
    \caption{A comparative evaluation of our method against state-of-the-art approaches using ranking correlation metrics—Spearman’s $\rho$ and Kendall’s $\tau$. Methods in the unsupervised or self-supervised category are highlighted in gray. Columns \textbf{SUP} and \textbf{SSL} indicate whether a method is supervised or self-supervised, respectively; \textbf{FT} denotes whether fine-tuning with SSL was applied; \textbf{ATTN} indicates the use of attention mechanisms; and \textbf{GFLOPs} reflects the computational complexity of each method.}
    \label{tab:sota_comparison}
    \begin{tabular}{lcc|cc|c|c|c|c|c}
        \toprule
        \multirow{2}{*}{Method} & \multicolumn{2}{c|}{TVSum} & \multicolumn{2}{c}{SumMe} & SUP & SSL & FT & ATTN & GFLOPs \\
        & $\tau$ & $\rho$ & $\tau$ & $\rho$ \\
        \midrule
        Human & 0.177 & 0.204 & 0.212 & 0.212 & - & - & - & - & -\\
        Random & 0.000 & 0.000 & 0.000 & 0.000 & - & - & - & - & - \\
        \midrule
        dppLSTM \cite{dpplstm} & 0.042 & 0.055 & -0.026 & -0.031 & \textbf{\checkmark} & \xmark & \xmark & \xmark & - \\
        \rowcolor{gray!20}
        DR-DSN \cite{zhou2018deep} & 0.020 & 0.026 & 0.043 & 0.050 & \xmark & \xmark & \xmark & \xmark & - \\
        DR-DSN$_{2000}$\cite{zhou2018deep} & 0.152 & 0.198 & -0.016 & -0.022 & - & - & - & \xmark & - \\
        \rowcolor{gray!20}
        GANSUM \cite{gansum} & 0.024 & 0.032 & -0.010 & -0.012 & \xmark & \xmark & \xmark & \xmark & - \\
        \rowcolor{gray!20}
        CSNet\cite{csnet} & 0.025 & 0.034 & - & - & \xmark & \xmark & \xmark & \textbf{\checkmark} & - \\
        \rowcolor{gray!20}
        GANSUM+AAE\cite{gansum_aae} & -0.047 & -0.062 & -0.018 & -0.023 & \xmark & \xmark & \xmark & \textbf{\checkmark} & -\\
        DAC\cite{dac} & 0.058 & 0.065 & 0.063 & 0.059 & \textbf{\checkmark} & \xmark & \xmark & \textbf{\checkmark} & - \\
        \rowcolor{gray!20}
        GANSUM+GL+RPE \cite{gansum_gl_rpe} & 0.064 & 0.084 & - & -  & \xmark & \xmark & \xmark & \xmark & - \\
        \rowcolor{gray!20}
        CSNet+GL+RPE\cite{gansum_gl_rpe} & 0.070 & 0.091 & - & -  & \xmark & \xmark & \xmark & \xmark & -\\
        HSA-RNN\cite{hsa} & 0.082 & 0.088 & 0.064 & 0.066 & \textbf{\checkmark} & \xmark & \xmark & \xmark & -\\
        DAN\cite{dan} & 0.071 & 0.099 & - & - & \textbf{\checkmark} & \xmark & \xmark & \textbf{\checkmark} & -\\
        HMT\cite{hmt} & 0.096 & 0.107 & 0.079 & 0.080 & \textbf{\checkmark} & \xmark & \xmark & \textbf{\checkmark} & - \\
        VJMHT\cite{vjmht} & 0.097 & 0.105 & 0.106 & 0.108 & \textbf{\checkmark} & \xmark & \xmark & \textbf{\checkmark} & 56.5G \\
        STVT\cite{stvt} & 0.1 & 0.131 & - & - & \textbf{\checkmark} & \xmark & \xmark & \textbf{\checkmark} & -\\
        DSNet-AB\cite{dsnet} & 0.108 & 0.129 & 0.051 & 0.059 & \textbf{\checkmark} & \xmark & \xmark & \textbf{\checkmark} & 4.14G \\
        DSNet-AF\cite{dsnet} & 0.113 & 0.138 & - & - & \textbf{\checkmark} & \xmark & \xmark & \textbf{\checkmark} & 3.8G \\
        \rowcolor{gray!20}
        MA \cite{ma} & 0.110 & 0.144 & 0.046 & 0.057 & \xmark  & \textbf{\checkmark} & \textbf{\checkmark} & \textbf{\checkmark} & - \\
        CLIP-It\cite{clipit} & 0.108 & 0.147 & - & - & \textbf{\checkmark} & \xmark & \xmark & \textbf{\checkmark} & -\\
        iPTNet\cite{ipnet} & 0.134 & 0.163 & 0.101 & 0.119 & \textbf{\checkmark} & \xmark & \xmark & \textbf{\checkmark} & -\\
        A2Summ \cite{a2summ} & 0.137 & 0.165 & 0.108 & 0.129 & \textbf{\checkmark} & \xmark & \xmark & \textbf{\checkmark} & -\\
        VASNet \cite{vasnet} & 0.160 & 0.170 & 0.160 & 0.170 & \textbf{\checkmark} & \xmark & \xmark & \textbf{\checkmark} & 4.6G\\
        MAAM\cite{maam} & 0.179 & 0.236 & - & - & \textbf{\checkmark} & \xmark & \xmark & \textbf{\checkmark} & -\\
        VSS-Net \cite{vssnet} & 0.190 & 0.249 & - & - & \textbf{\checkmark} & \xmark & \xmark & \textbf{\checkmark} & -\\
        DMASum \cite{dmasum} & 0.203 & 0.267 & 0.063 & 0.089 & \textbf{\checkmark} & \xmark & \xmark & \textbf{\checkmark} & -\\
        RR-STG \cite{rrstg} & 0.162 & 0.212 & 0.211 & 0.234 & \textbf{\checkmark} & \xmark & \xmark & \textbf{\checkmark}  & - \\
        MSVA \cite{msva} & 0.190 & 0.210 & 0.200 & 0.230 & \textbf{\checkmark} & \xmark & \xmark & \textbf{\checkmark} & 11.62G \\
        SSPVS \cite{sspvs} & 0.181 & 0.238 & 0.192 & 0.257 & \textbf{\checkmark} & \textbf{\checkmark}  & \textbf{\checkmark} & \textbf{\checkmark} & 88.44G \\
        CSTA \cite{csta} & 0.194 & 0.255 & 0.246 & 0.274 & \textbf{\checkmark} & \xmark & \xmark & \textbf{\checkmark} & 31.46G \\
        \midrule
        \rowcolor{gray!20}
        \textbf{TRIM(Ours)} & \textbf{0.181} & \textbf{0.228} & \textbf{0.127} & \textbf{0.140} & \xmark & \textbf{\checkmark} & \textbf{\checkmark} & \xmark & \textbf{3.27G}\\
        \bottomrule
    \end{tabular}
\end{table}

\subsection{Training and Optimization}
\label{secc:train_optimize}
We pre-trained the proposed network for 60 epochs during stage 1 following the same optimization protocol from Zhou et al. \cite{zhou2018deep}. For stage 2 unsupervised fine-tuning, we adopted the optimization protocol from Zhou et al. \cite{zhou2018deep} with minor adjustments. Our proposed model was trained using the Adam optimizer \cite{adam_optimizer}, with a learning rate of \(1 \times 10^{-5}\) and weight decay of \(1 \times 10^{-5}\) for 90 epochs. 

\subsection{Evaluation Methods}
\label{secc:eval_methods}
We follow a strict evaluation protocol same as \cite{csta,maam} by running each experiment 10 times where each experiment follows a five-fold cross-validation to reflect all videos within the test set. We assess the performance of our proposed method by computing rank-based correlation coefficients—specifically, Kendall’s $\tau$ and Spearman’s $\rho$—between the predicted outputs and the ground truth motivated by recent SOTA approaches in video summarization [\cite{csta,maam,vssnet,sspvs,ipnet,clipit}]. The average performance across 10 independent runs is reported as the final performance of our model throughout this paper. Early approaches to video summarization primarily relied on F1 score as their sole evaluation protocol. However, due to the constraint on summary length, the F1 score can be artificially inflated when the model favors selecting many short shots over fewer longer ones, as discussed in \cite{rethinking,maam,csta}. For instance, DRDSN \cite{zhou2018deep} reports an F1 score of 57.6 on the TVSUM \cite{tvsum_dataset} dataset, despite exhibiting a low correlation with human-annotated ground truth, as shown in Table~\ref{tab:sota_comparison}. To overcome this limitation, we evaluate rank correlation coefficients—consistent with existing state-of-the-art (SOTA) methods—and present a comparative analysis in the following section.

TRIM adopts the same methodology as \cite{csta} to generate video summaries. During inference, we first compute \( p_t \) for each video \( V \) and apply Kernel Temporal Segmentation (KTS) \cite{kts} to divide the video into shots and calculate the mean score for each shot. Finally, we employ the Knapsack algorithm to select a subset of shots such that the total summary length does not exceed 15\% of the original video duration following \cite{tvsum_dataset}, prioritizing shots based on their mean scores and lengths.

\begin{figure}[htbp]
    \centering
    \begin{subfigure}[b]{0.49\textwidth}
        \centering
        \includegraphics[width=\textwidth]{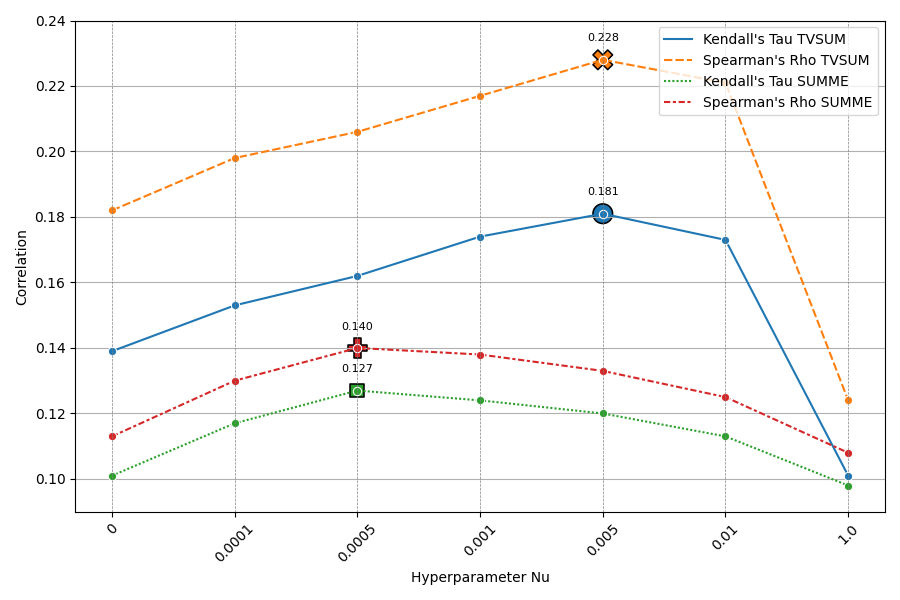}
        \caption{Sensitivity of $\nu$ on SUMME and TVSUM datasets.}
        \label{fig:hyperparam_nu}
    \end{subfigure}
    \hfill
    \begin{subfigure}[b]{0.49\textwidth}
        \centering
        \includegraphics[width=\textwidth]{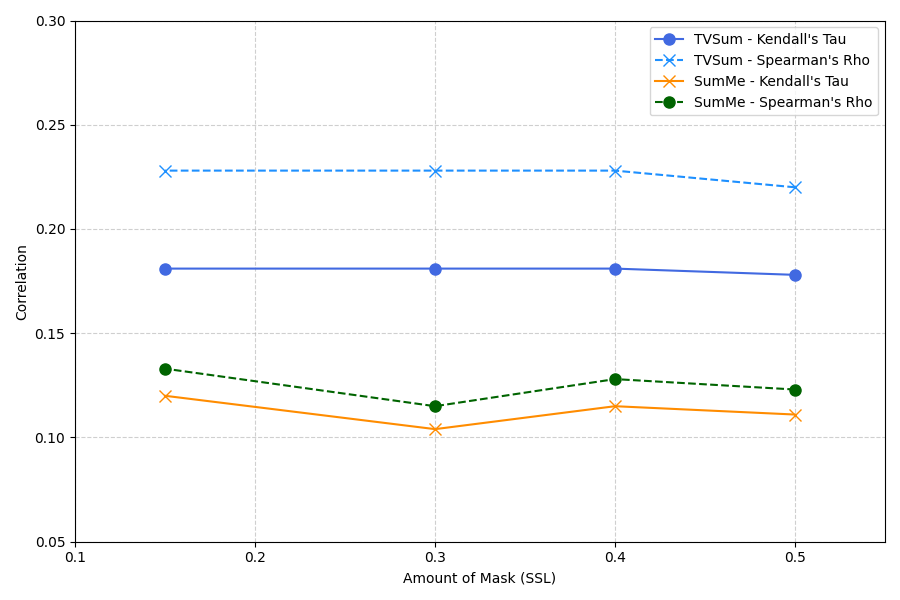}
        \caption{Ablation study on mask ratio during SSL.}
        \label{fig:mask_ssl}
    \end{subfigure}
    \caption{Ablation of mask ratio $m$ and hyperparameter $\nu$. Fig.~\ref{fig:hyperparam_nu} (left): Sensitivity of hyperparameter $\nu$ within $L_{PRE}$ loss function during two-stage training. Fig.~\ref{fig:mask_ssl} (right): Performance of proposed model when pre-trained with SSL at different mask ratio. The mask ratio could be between $0$ to $0.5$, where $0.5$ represents that half of the frames in the video have been randomly masked, and this has been done randomly for both views during SSL. Here $\alpha=1,\beta=5,\gamma=0$ for TVSUM~\cite{tvsum_dataset} dataset and $\alpha=1,\beta=0, \gamma=0.5$ for SUMME~\cite{summe_dataset} dataset.}
    \label{fig:nu_mask_ratio}
\end{figure}

\begin{figure}[htbp]
    \centering
    \begin{subfigure}[b]{0.49\textwidth}
        \centering
        \includegraphics[width=\textwidth]{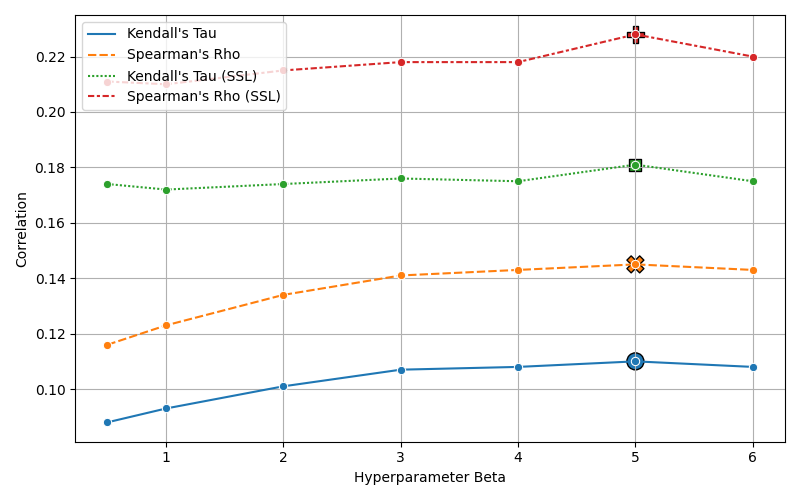}
        \caption{ Sensitivity of $\beta$ on TVSUM\cite{tvsum_dataset} dataset}
        \label{fig:subfig1}
    \end{subfigure}
    \hfill
    \begin{subfigure}[b]{0.49\textwidth}
        \centering
        \includegraphics[width=\textwidth]{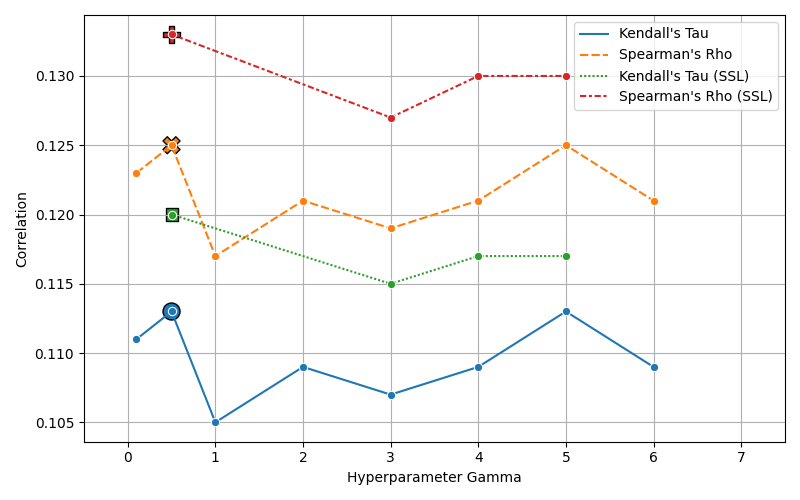}
        \caption{Sensitivity of $\gamma$ on SUMME\cite{summe_dataset} dataset.}
        \label{fig:subfig2}
    \end{subfigure}
    \caption{Hyper-parameter sensitivity analysis for $\beta$ and $\gamma$ parameters. Fig.~\ref{fig:subfig1} (left): Performance of the proposed model for different values of $\beta$, while $\alpha = 1, \gamma=0$ in our proposed loss function $L_\text{UNSUP}$ in \eqref{eq:mainloss}, with and without SSL Pre-training. Fig.~\ref{fig:subfig2} (right): Performance of the proposed model for different values of $\gamma$, while $\alpha = 1, \beta=0$ in our proposed loss function $L_\text{UNSUP}$, with and without SSL pre-training. }
    \label{fig:hyp_beta}
\end{figure}

\subsection{Ablation Studies}
\label{ssec:ablations}
This section presents an evaluation of the contributions of each component in our losses \( L_{\text{PRE}} \) and \( L_{\text{UNSUP}} \). The ablation study on \( L_{\text{UNSUP}} \) that we present on Table~\ref{tab:ablation} indicates that the optimal performance is achieved with hyperparameter values \( \alpha = 1 \), \( \beta = 5 \), and \( \gamma = 0 \), on TVSUM~\cite{tvsum_dataset} dataset and \( \alpha = 1 \), \( \beta = 0 \), and \( \gamma = 0.5 \) for SUMME~\cite{summe_dataset} dataset, respectively. 
An intuitive rationale for hyperparameter selection is detailed in \ref{secc:hyperparameter_rationale}.
A comprehensive ablation analysis examining the individual contributions of each loss component to the alignment between the predicted outputs and the ground truth is provided in Sect.~\ref{secc:qualitative_analysis} in the Appendix. Furthermore, we provide evidence that our network does not produce random importance score which has been highlighted in ~\ref{secc:qualitative_analysis}.


\begin{table}[h]
    \centering
    \small
    \caption{
Ablation study of the individual loss components in Eq.~\eqref{eq:mainloss} on the SumMe~\cite{summe_dataset} and TVSum~\cite{tvsum_dataset} benchmarks. We report results for both the single-stage unsupervised setting (Stage~2 only) and the full two-stage training framework. This comparison highlights the contribution of each loss term and demonstrates how their combination improves summarization performance.
}
    \begin{tabular}{lcc|cc}
        \toprule
        \textbf{Loss} & \multicolumn{2}{c}{\textbf{TVSum}} & \multicolumn{2}{c}{\textbf{SumMe}} \\
        & $\tau$ & $\rho$ & $\tau$ & $\rho$ \\
        \midrule
        Unsupervised method (Stage 2 only) \\
        \midrule
        $L_\text{PTRIM} + L_\text{SD}$ & 0.1015 & 0.1336 & 0.1094 & 0.1217 \\
        $L_\text{PCTRIM} + L_\text{SD}$ & 0.0604 & 0.0796 & 0.1134 & 0.1261 \\
        $L_\text{REP} + L_\text{SD}$ & 0.0740 & 0.0970 & 0.1085 & 0.1207 \\
        $\alpha L_\text{PTRIM} + \gamma L_\text{PCTRIM}  + L_\text{SD}$ & 0.0948 & 0.1248 & 0.1155 & 0.1275 \\
        $\alpha L_\text{PTRIM} + \beta L_\text{REP} + L_\text{SD}$ & 0.1100 & 0.1452 & 0.1129 & 0.1254 \\
        $\alpha L_\text{PTRIM} + \beta L_\text{REP} +  \gamma  L_\text{PCTRIM} + L_\text{SD}$ & 0.1100 & 0.1452 & 0.1155 & 0.1275  \\
        \midrule
        Self-Supervised method (Both stages)\\
        \midrule
        $L_\text{PTRIM} + L_\text{SD}$ & 0.1830 & 0.2181 & 0.1170 & 0.1300 \\
        $L_\text{PCTRIM} + L_\text{SD}$ & 0.1785 & 0.2138 & 0.1139 & 0.1265 \\
        $L_\text{REP} + L_\text{SD}$ & 0.1693 & 0.2193 & 0.1005 & 0.1115 \\
        $\alpha L_\text{PTRIM} + \gamma L_\text{PCTRIM}  + L_\text{SD}$ & 0.1846 & 0.2194 & 0.1268 & 0.1408\\
        $\alpha L_\text{PTRIM} + \beta L_\text{REP} + L_\text{SD}$ & 0.1810 & 0.2282 & 0.0981 & 0.1088 \\
        $\alpha L_\text{PTRIM} + \beta L_\text{REP} +  \gamma  L_\text{PCTRIM} + L_\text{SD}$ & 0.1810 & 0.2274 & 0.1268 & 0.1408  \\
        \bottomrule
    \end{tabular}
    \label{tab:ablation}
\end{table}

Additionally, we conduct an ablation study to investigate the impact of masking ratio during the first stage (self-supervised pre-training) of our proposed method. The results are  illustrated in Fig.~\ref{fig:mask_ssl} and indicate that applying random masking to approximately 15--30\% of video frames is sufficient to achieve SOTA performance on both datasets. 
Moreover, we have studied the sensitivity of hyper-parameter $\nu$ in Fig.~\ref{fig:hyperparam_nu} and found that 0.005 obtains the best results for TVSUM~\cite{tvsum_dataset} dataset while 0.0005 obtains the best results on SUMME~\cite{summe_dataset} dataset during SSL pre-training.

We have also included the ablation studies on hyper-parameter $\beta$ on TVSUM dataset and hyper-parameter $\gamma$ on SUMME dataset and our results conclusively indicate $\beta = 5$ and $\gamma = 0.5$ achieves the SOTA results on our proposed network. Our conclusions have been aligned on both unsupervised and two-stage method with SSL as illustrated in Fig.~\ref{fig:hyp_beta}. We observed that during unsupervised phase $\gamma =0.5$ and $\gamma =5$ both achieves SOTA results however, two stage method conclusively proves $\gamma =0.5$ is the optimal value for the hyper-parameter on SUMME dataset.

\paragraph{5. Conclusions}
This papers introduces a two-stage self-supervised method for summarizing videos that achieves state-of-the-art results on TVSUM \cite{tvsum_dataset} and SUMME \cite{summe_dataset} datasets with minimal computational overhead. Unlike supervised methods, the proposed metrics and loss functions are inherently general and operate independently of any dataset-specific spatio-temporal relationships or annotations. Thus, the proposed method can generalize better across datasets. The proposed metrics and network can also be widely adopted to video surveillance and anomaly detection tasks due to their computational efficiency.

\bibliography{references}

\newpage
\appendix

\section{Appendix / supplemental material}

\subsection{Supplementary Evaluation Results}
In this section, we further provide additional performance in terms of F1 score similar to early stage video summarization methods in tandem with ranked correlation metrics presented earlier. The result is shown in table \ref{tab:F1_score}. This is to be noted that, the performance of our proposed method reported in Table \ref{tab:F1_score} follows a strict evaluation protocol towards running every experiment 10 times and reporting the mean performance. In comparison, some early stage video summarization methods only reported the best score from a single experiment.
\label{ssec:additional_results} 

To ensure that our model's performance is not due to random outputs, we conducted two control experiments. In the first experiment, we bypassed training and directly performed inference using the untrained network to generate random importance scores. These scores were then evaluated against the ground truth using rank correlation metrics. As with all other experiments, we repeated this process 10 times and reported the mean correlation scores. The results show no correlation between the random predictions and the ground truth. To further validate this finding, we conducted a second experiment in which the model was trained solely with the \( L_{\text{SD}} \) loss, which encourages higher variance in importance scores but does not capture any meaningful learning objective. This setup also resulted in near-zero correlation with the ground truth, reinforcing the conclusion that meaningful loss design is necessary for achieving high correlation with ground truth. The results of both experiments are summarized in Table~\ref{tab:random_results}.

Furthermore, we conducted an additional ablation study to examine the impact of two different feature extraction strategies. In the first approach, all video frames were resized directly to \(224 \times 224\). In the second approach, we followed the pre-processing method from \cite{ma}, where frames were first resized to \(256 \times 256\) and then center-cropped to \(224 \times 224\). Our findings indicate that the second method enhances performance in the unsupervised setting (stage 2) and provides a marginal improvement when applied to our proposed two-stage method. The results of these experiments are summarized in Table~\ref{tab:performance}.

\begin{table}[ht]
    \centering
    \caption{A comparison of our proposed method against state-of-the-art approaches in terms of F1 score on the benchmark video summarization datasets SumMe \cite{summe_dataset} and TVSum \cite{tvsum_dataset}. The first row contains unsupervised methods, whereas the second row includes supervised ones.}
    \begin{tabular}{lcccccc}
    \toprule
        Method & CNN & GAN & RNN  & Attention & SumMe & TVSUM \\
    \midrule
    \rowcolor{gray!20}
    SGAN \cite{gansum} & \xmark & \textbf{\checkmark} & \xmark  & \xmark & 38.7 & 50.8 \\
    \rowcolor{gray!20}
    DRSN~\cite{zhou2018deep} & \xmark & \xmark & \textbf{\checkmark}  & \xmark & 41.1 & 57.6 \\
    \rowcolor{gray!20}

    GAN\_DPP \cite{gansum} & \xmark & \textbf{\checkmark} & \xmark  & \xmark & 39.1 & 51.7 \\
    \rowcolor{gray!20}
    ACGAN \cite{acgan} & \xmark & \textbf{\checkmark} & \xmark  & \xmark & 46.0 & 58.5 \\
    \rowcolor{gray!20}
    WS-HRL \cite{wshrl} & \xmark & \xmark & \textbf{\checkmark}  & \xmark & 43.6 & 58.4 \\
    \midrule
    DRDSN\_SUP~\cite{zhou2018deep} & \xmark & \xmark & \textbf{\checkmark}  & \xmark & 42.1 & 58.1 \\
    vsLSTM \cite{dpplstm} & \xmark & \xmark & \textbf{\checkmark}  & \xmark & 37.6 & 54.2 \\
    SGAN\textsubscript{S} \cite{gansum} & \xmark  & \textbf{\checkmark} & \xmark  & \xmark  & 41.7 & 56.3 \\
    H-RNN \cite{hrnn} & \xmark & \xmark & \textbf{\checkmark}  & \xmark & 42.1 & 57.9 \\
    HSA \cite{hsa} & \xmark & \xmark & \textbf{\checkmark}  & \xmark & 42.3 & 58.7 \\
    S-FCN \cite{sfcn} & \textbf{\checkmark} & \xmark & \xmark  & \xmark & 47.5 & 56.8 \\
    CSNET\textsubscript{S} \cite{dsnet} & \textbf{\checkmark} & \xmark & \xmark  & \xmark & 48.6 & 58.5 \\
    SSPVS \cite{sspvs} & \xmark & \xmark & \xmark  & \textbf{\checkmark} & 48.7 & 60.3 \\
    \midrule
    \rowcolor{gray!20}
    \textbf{TRIM(Ours)} & \textbf{\checkmark} & \xmark & \xmark  & \xmark & \textbf{43.67} & \textbf{59.61} \\
    \bottomrule

    \end{tabular}
    \label{tab:F1_score}
\end{table}

\begin{table}[h]
    \centering
    \caption{Ablation analysis conducted on the TVSum\cite{tvsum_dataset} and SumMe\cite{summe_dataset} datasets, assessing the impact of resizing and center crop augmentations on ranking correlation coefficients for the proposed self-supervised and unsupervised frameworks.}
    \begin{tabular}{lc|cc|cc}
        \toprule
        \textbf{Method} & \textbf{SSL} & \multicolumn{2}{c|}{\textbf{TVSum}} & \multicolumn{2}{c}{\textbf{SumMe}} \\
        & & $\tau$ & $\rho$ & $\tau$ & $\rho$ \\
        \midrule
        Baseline $224 \times 224$ resize & \xmark & 0.110 & 0.145 & 0.115 & 0.127 \\
        Resize 256 Center Crop $224 \times 224$ & \xmark & 0.146 & 0.192 & 0.118 & 0.132 \\
        Baseline $224 \times 224$ resize & \checkmark & 0.174 & 0.222 & 0.125 & 0.136 \\
        Resize 256 Center Crop $224 \times 224$ & \checkmark & 0.181 & 0.228 & 0.127 & 0.140 \\
        \bottomrule
    \end{tabular}
    \label{tab:performance}
\end{table}

\begin{table}[h]
    \centering
    \caption{Experimental evidence indicating that the proposed model, when subjected to either no training or optimization using a random loss function, produces outputs that exhibit no statistically significant correlation with the ground truth. These results underscore the critical role of our proposed loss function design and proper training in achieving meaningful model performance.}
    \begin{tabular}{lcc}
        \toprule
        \textbf{Loss} & \multicolumn{2}{c}{\textbf{TVSum}} \\
        & $\tau$ & $\rho$ \\
        \midrule
        Evaluation of Random Experiments on Network \\
        \midrule
        Random Initialization(no training) & -0.0015 & -0.0018 \\
        Trained with $L_\text{SD}$ & -0.0122 & -0.0160 \\
        \midrule
        \bottomrule
    \end{tabular}
    \label{tab:random_results}
\end{table}

\subsection{Additional Quantitative Analysis}
In this section, we analyze the alignment between the predicted importance score by our proposed network on unsupervised method(Stage 2 only) and self-supervised method with unsupervised fine-tuning(two-stage method) compared to mean ground truth annotations by 20 users in TVSUM\cite{tvsum_dataset} dataset.
\label{secc:qualitative_analysis}
\subsubsection{Unsupervised Method(Stage 2)}
\label{secc:qualitative_analysis_unsup}
\begin{figure}[H]
    \centering

    \begin{subfigure}[t]{0.49\textwidth}
        \centering
        \includegraphics[width=\linewidth]{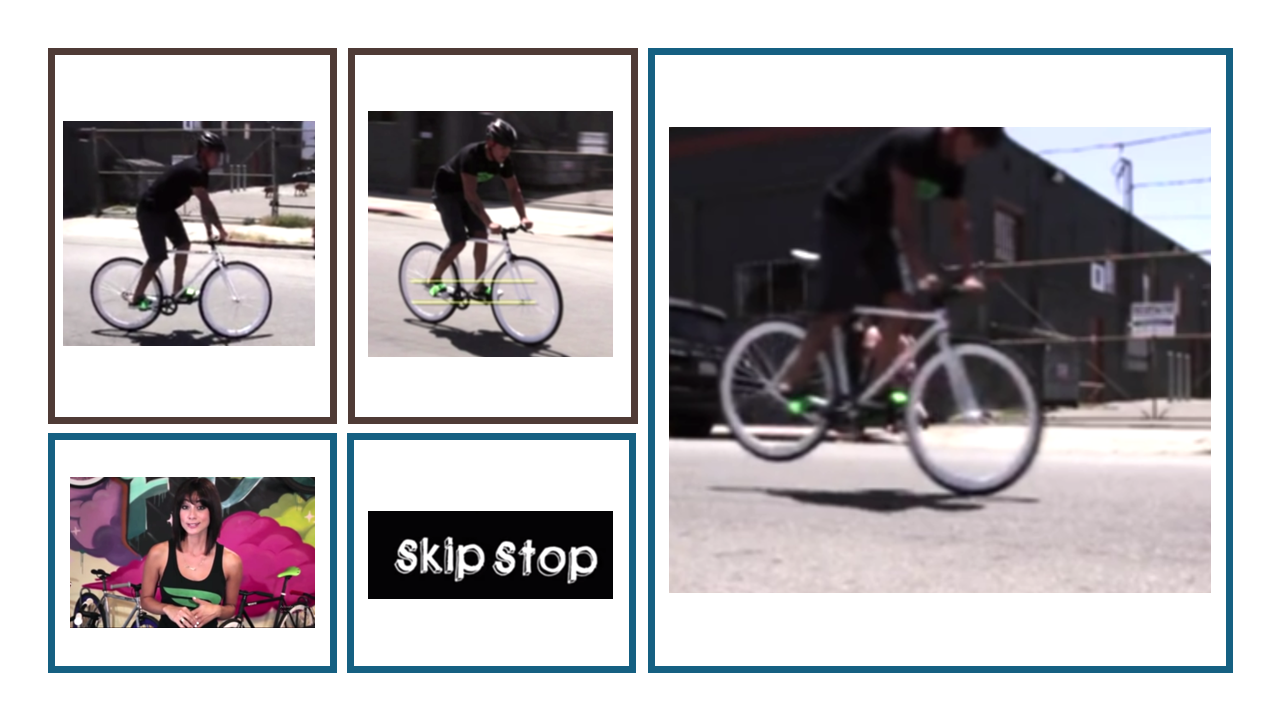}
        \caption{Key Frames from Video "eQu1rNs0an0"}
    \end{subfigure}
    \begin{subfigure}[t]{0.49\textwidth}
        \centering
        \includegraphics[width=\linewidth]{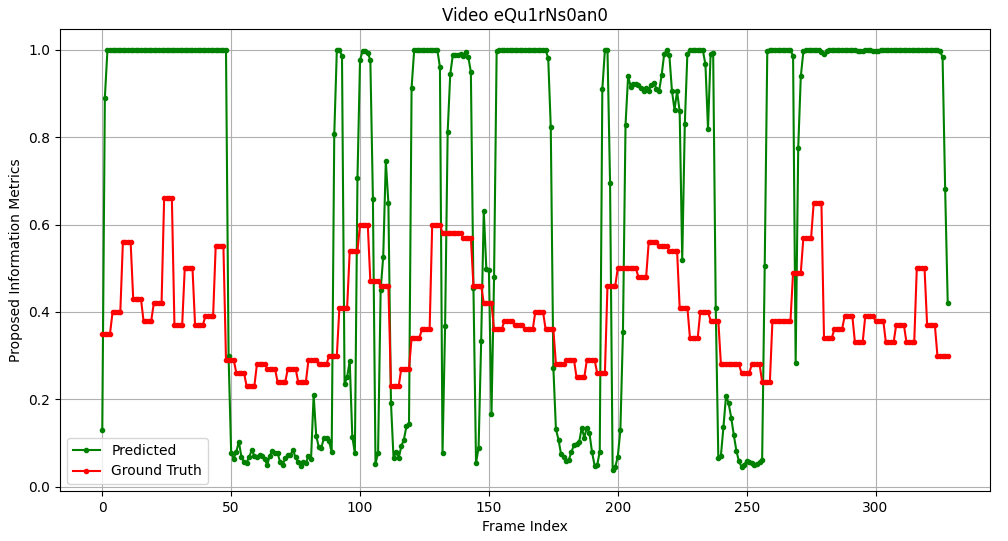}
        \caption{$L_\text{PTRIM} + L_\text{SD}$}
    \end{subfigure}

    \vspace{0.3cm}

    \begin{subfigure}[t]{0.49\textwidth}
        \centering
        \includegraphics[width=\linewidth]{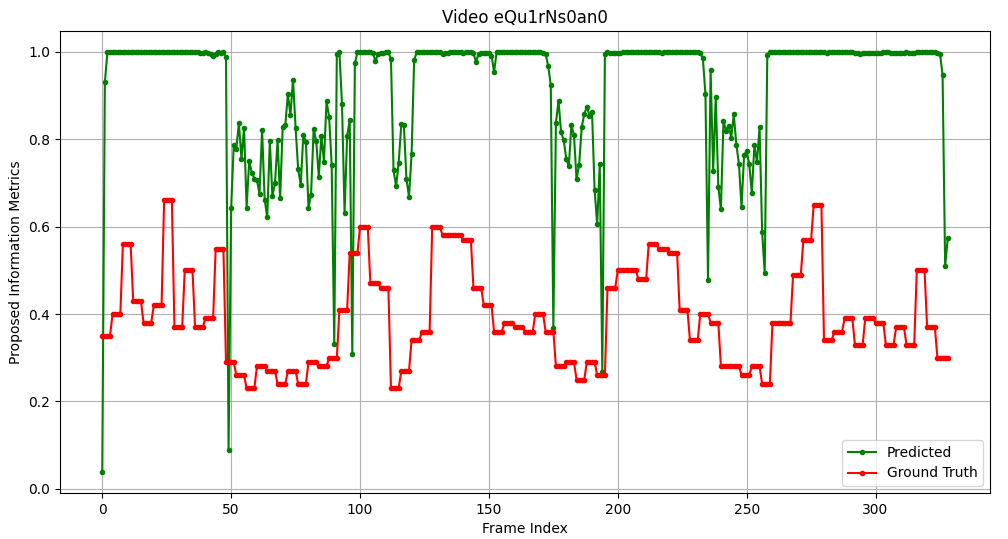}
        \caption{$L_\text{REP} + L_\text{SD}$}
    \end{subfigure}
    \begin{subfigure}[t]{0.49\textwidth}
        \centering
        \includegraphics[width=\linewidth]{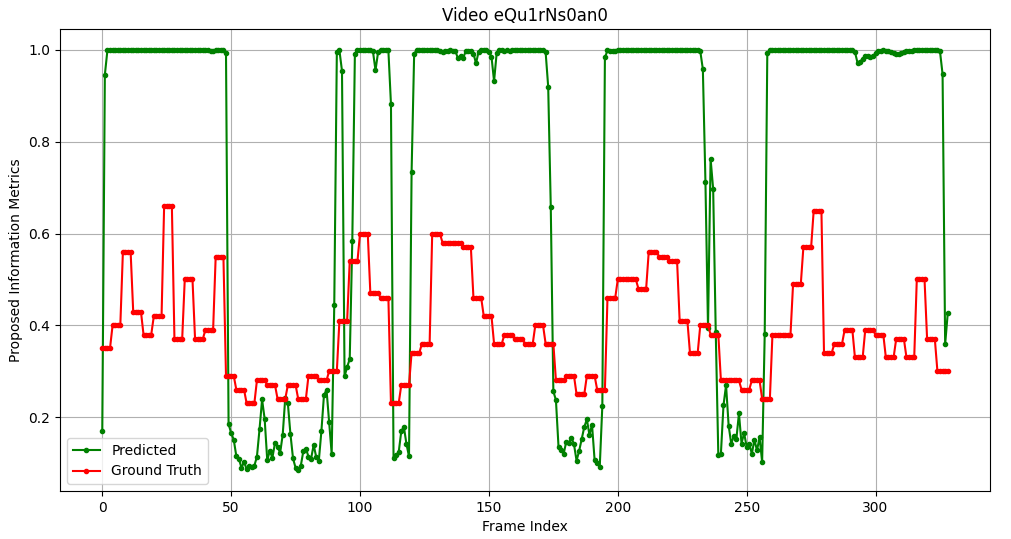}
        \caption{$L_\text{PCTRIM} + L_\text{SD}$}
    \end{subfigure}

    \vspace{0.3cm}

    \begin{subfigure}[t]{0.49\textwidth}
        \centering
        \includegraphics[width=\linewidth]{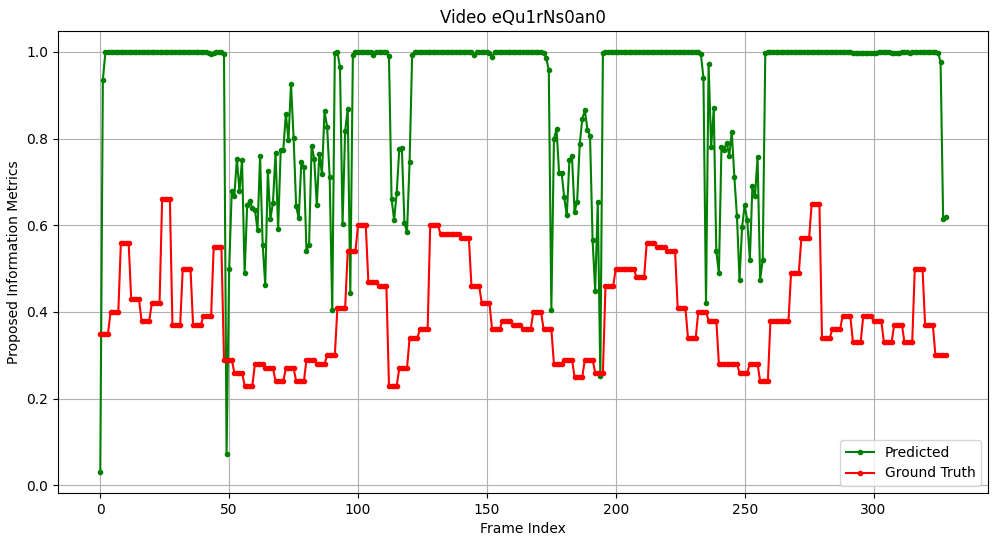}
        \caption{$\alpha L_\text{PTRIM} + \beta L_\text{REP} + L_\text{SD}$}
    \end{subfigure}
    \begin{subfigure}[t]{0.49\textwidth}
        \centering
        \includegraphics[width=\linewidth]{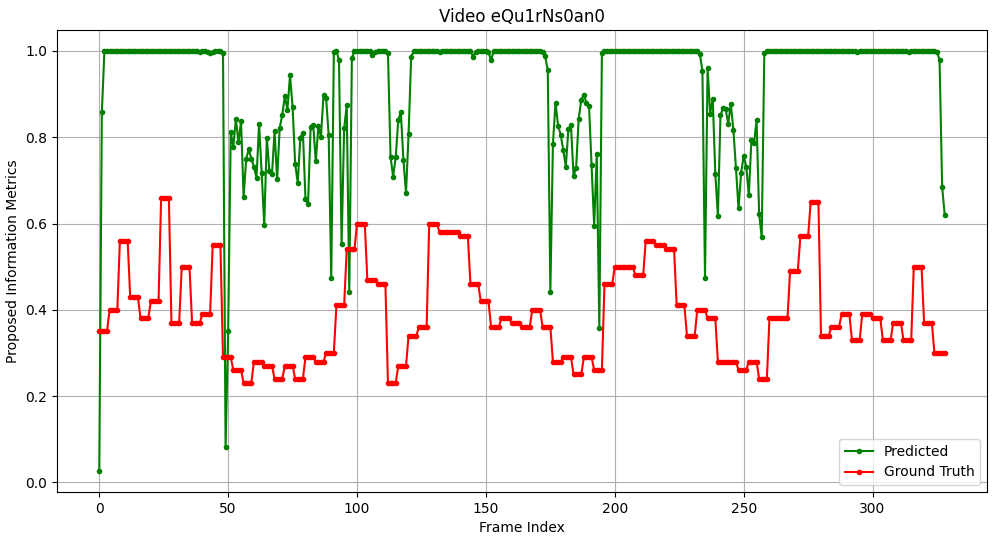}
        \caption{$\alpha L_\text{PTRIM} +  \beta L_\text{REP} + \gamma L_\text{PCTRIM} + L_\text{SD}$}
    \end{subfigure}

    \caption{Ablation study[from Table \ref{tab:ablation}] evaluating the contribution of each loss component, based on the comparison between predicted importance scores and the ground-truth annotations (averaged across 20 annotators) on video "eQu1rNs0an0" within the TVSum\cite{tvsum_dataset} dataset.}
    \label{fig:unsup_eQu1rNs0an0}
\end{figure}

The visual interpretation of predicted vs ground truth importance score for stage 2 is outlined as follows.
 we can observe more frames were dropped from selection when $L_\text{PTRIM}$ loss was trained as shown in Fig.~\ref{fig:unsup_eQu1rNs0an0}(b). This aligns with our intuition that $L_\text{PTRIM} + L_\text{SD}$ should only choose frames which are semantically diverse in nature. On the other hand, when our network was trained with $L_\text{REP} + L_\text{SD}$ most of the representative frames were selected as shown in \ref{fig:unsup_eQu1rNs0an0}(c). When both losses were combined as $\alpha L_\text{PTRIM} + \beta L_\text{REP} + L_\text{SD}$ we observe that some of the frames that were discarded by $L_\text{PTRIM} + L_\text{SD}$ due to low importance are much higher since $L_\text{PTRIM}$ and $L_\text{REP}$ compliments each other in the same way as described by \cite{zhou2018deep} in terms of diversity and representativeness to summarize a video in an unsupervised scenario. We present this in Fig.~\ref{fig:unsup_eQu1rNs0an0}(e).

On the other hand, $L_\text{PCTRIM} + L_\text{SD}$ works in a different way than all other losses, its history-dependent structure captures diversity from past information signals on videos with shorter length since the context does not change dramatically as seen in SUMME \cite{summe_dataset} dataset. Hence the behavior is different from $L_\text{PTRIM}$ and $L_\text{REP}$ losses. This can be observed in Fig~\ref{fig:unsup_eQu1rNs0an0}(d). The combined behavior of all losses $\alpha L_\text{PTRIM} +  \beta L_\text{REP} + \gamma L_\text{PCTRIM} + L_\text{SD}$ [Fig.~\ref{fig:unsup_eQu1rNs0an0}(f)] seems similar to $\alpha L_\text{PTRIM} +  \beta L_\text{REP} + L_\text{SD}$ since $\gamma=0.5$ or $\gamma=0$ produce the same result for TVSUM dataset.

\subsubsection{Self-Supervised Pre-training with Unsupervised fine-tuning(Two-stage Method)}
The visual interpretation of predicted vs ground truth importance score for two stage method is outlined as follows. 

\textbf{Intuition and Advantages:} Our intuition behind using self-supervised pre-training was to train our proposed network that learns to predict importance score invariant to noise within the video data. Specifically, if we mask m\% percentage of frames on both augmentations randomly, the model would be able to align prediction scores between both views if some of the frames in the video appears trivial to learn the representation between them. Therefore, the pre-training stage will help the proposed network to learn importance between the frames within the video. Another advantage of self-supervised pre-training is reproducibility in second stage, where all networks in five-fold cross validation during stage 2 initializes with same weights for the proposed network. As a result, the standard deviation across the folds or the seeds have been minimal compared to random initialization. A detailed explanation with results is provided in Sect.~\ref{secc:reproducibility} of this Appendix.

\begin{figure}[H]
    \centering

    \begin{subfigure}[t]{0.49\textwidth}
        \centering
        \includegraphics[width=\linewidth]{qualitative_analysis/video_43_image.png}
        \caption{Key Frames from Video "eQu1rNs0an0"}
    \end{subfigure}
    \begin{subfigure}[t]{0.49\textwidth}
        \centering
        \includegraphics[width=\linewidth]{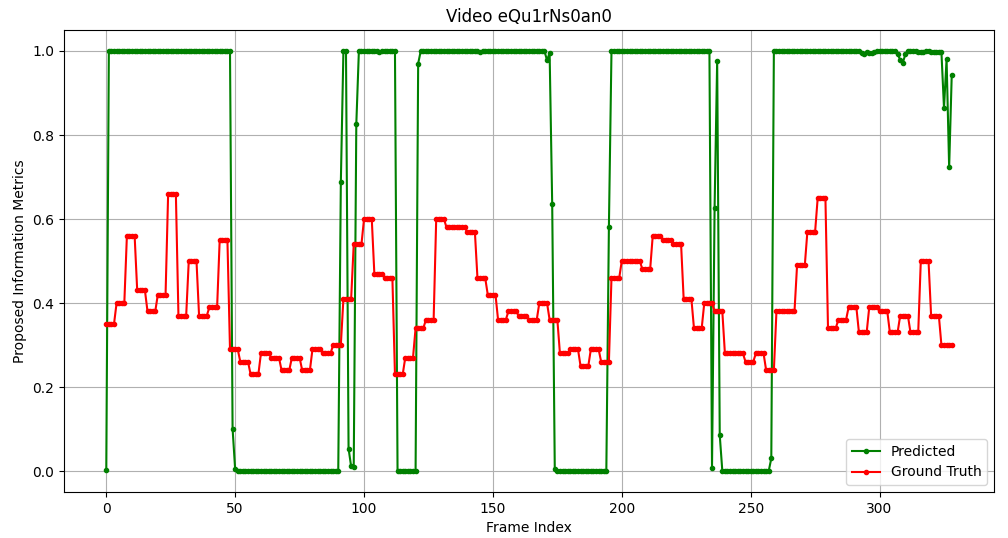}
        \caption{$L_\text{PTRIM} + L_\text{SD}$}
    \end{subfigure}

    \vspace{0.3cm}

    \begin{subfigure}[t]{0.49\textwidth}
        \centering
        \includegraphics[width=\linewidth]{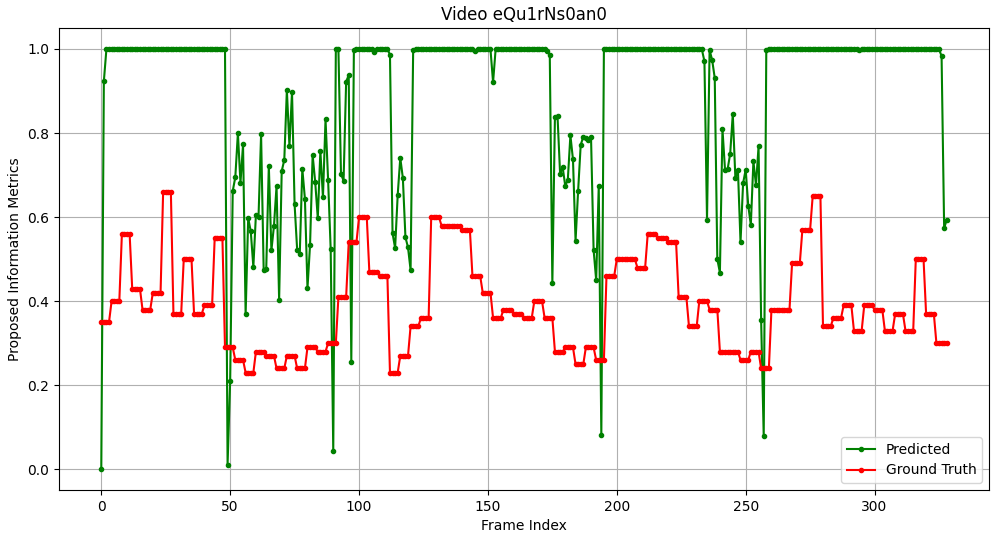}
        \caption{$L_\text{REP} + L_\text{SD}$}
    \end{subfigure}
    \begin{subfigure}[t]{0.49\textwidth}
        \centering
        \includegraphics[width=\linewidth]{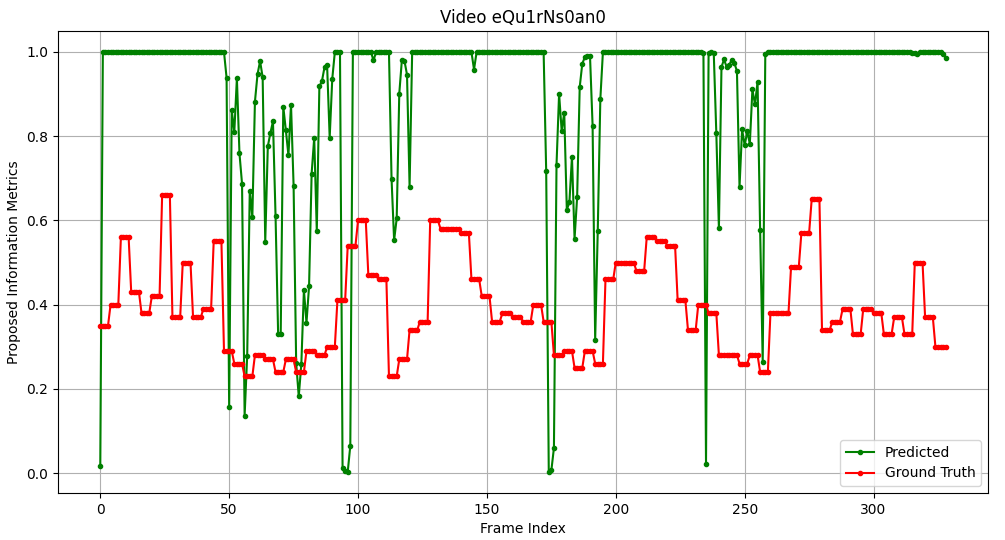}
        \caption{$L_\text{PCTRIM} + L_\text{SD}$}
    \end{subfigure}

    \vspace{0.3cm}

    \begin{subfigure}[t]{0.49\textwidth}
        \centering
        \includegraphics[width=\linewidth]{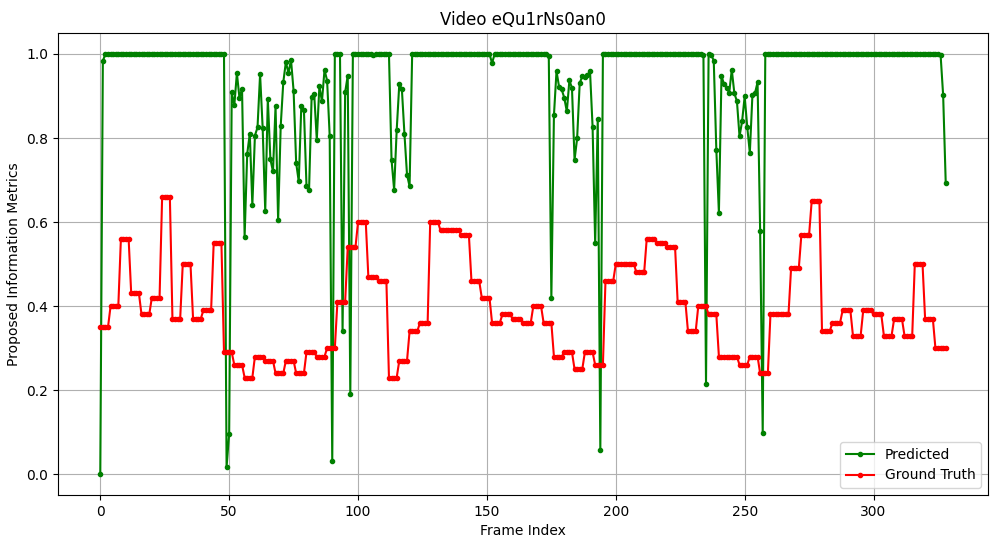}
        \caption{$\alpha L_\text{PTRIM} +  \beta L_\text{REP}  + L_\text{SD}$}
    \end{subfigure}
    \begin{subfigure}[t]{0.49\textwidth}
        \centering
        \includegraphics[width=\linewidth]{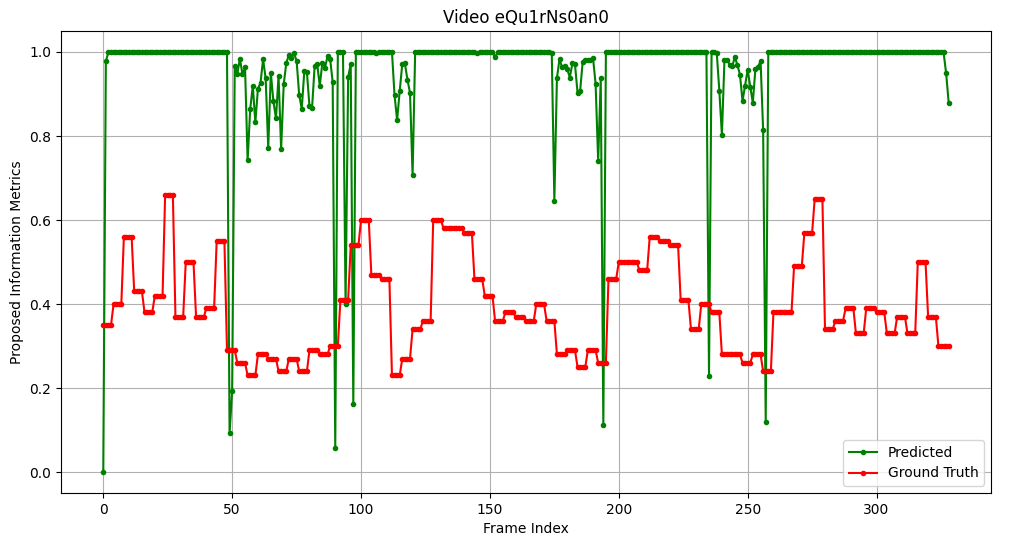}
        \caption{$\alpha L_\text{PTRIM} +  \beta L_\text{REP} + \gamma L_\text{PCTRIM} + L_\text{SD}$}
    \end{subfigure}

    \caption{Ablation study [from Table \ref{tab:ablation}] evaluating the contribution of each loss component, based on the comparison between predicted importance scores and the ground-truth annotations (averaged across 20 annotators) on video "eQu1rNs0an0" within the TVSum\cite{tvsum_dataset} dataset.}
    \label{fig:selfsup_eQu1rNs0an0}
\end{figure}

\textbf{Observation:} We can observe a similar behavior from the proposed unsupervised losses as discussed in Sect.~\ref{secc:qualitative_analysis_unsup} but with some subtle differences. In comparison to Fig.~\ref{fig:unsup_eQu1rNs0an0} we observe in Fig.~\ref{fig:selfsup_eQu1rNs0an0} the more important frames are close to 0.99 or 1 and less important frames are close to 0. This aligns with our intuition that self-supervised pre-training has learned the importance scores of the frames to maintain the representation of the video similar across both views with randomly masked frames. This pre-training phase helps the unsupervised fine-tuning further to achieve state-of-the-art results as presented in Table \ref{tab:ablation}.

\subsection{Dataset Characteristics and Hyperparameter Selection Rationale}

\subsubsection{Dataset Overview}
This study utilizes two standard video summarization datasets: SumMe \cite{summe_dataset} and TVSum \cite{tvsum_dataset}. TVSum contains 50 videos spanning 10 genres (e.g., documentaries, news, vlogs), each ranging from 2 to 10 minutes in length. In contrast, SumMe includes fewer videos with shorter durations of 1 to 6 minutes, featuring diverse content such as holidays, sports, and events, and captured with various camera styles including static, egocentric, and moving shots. While TVSum provides shot-level importance scores annotated by 20 users, SumMe offers frame-level selections averaged from at least 15 human-generated summaries.

\subsubsection{Hyperparameter Selection Rationale}
\label{secc:hyperparameter_rationale}
Since the two datasets differ significantly in the number of videos and the average length of each video, the choice of hyperparameters is tailored to suit their specific characteristics. For instance, our loss function $L_\text{PCTRIM}$ is more effective for shorter videos with fewer shots, making it well-suited for the SumMe \cite{summe_dataset} dataset. In contrast, $L_\text{REP}$ performs better on longer videos with multiple shots and numerous scene changes, which aligns better with the structure of TVSum \cite{tvsum_dataset}. Accordingly, we set $\beta = 0$ for SumMe, while for TVSum, both $\gamma = 0$ and $\gamma = 0.5$ yield comparable performance.

\subsection{Experiments Compute Resources}
\label{secc:exp_compute_resources}
The experimental computational resources for all experiments have been outlined in this section. The proposed model was trained on a single GPU NVIDIA TITAN RTX for all our experiments. The proposed model requires minimal computational requirements and can be ideal for resource-constrained environments.

\subsection{Reproducibility}
\label{secc:reproducibility}
Video summarization in general suffers from low reproducibility due to several stochasticity involved during training and prediction. The two most significant areas are network initialization across multiple splits in cross-validation and across random seeds. Also, the stochasticity involved during splitting videos randomly towards five fold cross-validation.

Many state-of-the-art methods train their models on each split or fold within five-fold cross-validation for varying numbers of epochs, optimizing performance individually for each fold. In contrast, our proposed method is consistently trained for the same number of epochs across all folds, ensuring better generalization across different data distributions. Specifically, we train Stage 1 (self-supervised learning) for 60 epochs and Stage 2 (unsupervised fine-tuning) for 90 epochs uniformly across all folds.

\begin{figure}[h!]
    \centering
    \includegraphics[width=\textwidth]{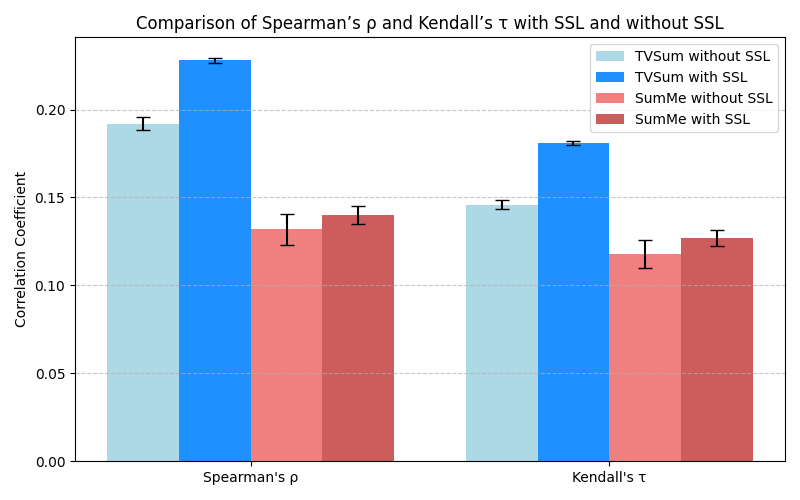}
    \caption{Comparison between mean and standard deviation of correlation coefficients across 10 experiments without SSL(stage 2 only) and with SSL(two stage proposed method) for SUMME\cite{summe_dataset} and TVSUM\cite{tvsum_dataset} datasets. Low standard deviation after SSL indicates better reproducibility.}
    \label{fig:reproducibilty}
\end{figure}

\begin{figure}[h!]
    \centering
    \includegraphics[width=\textwidth]{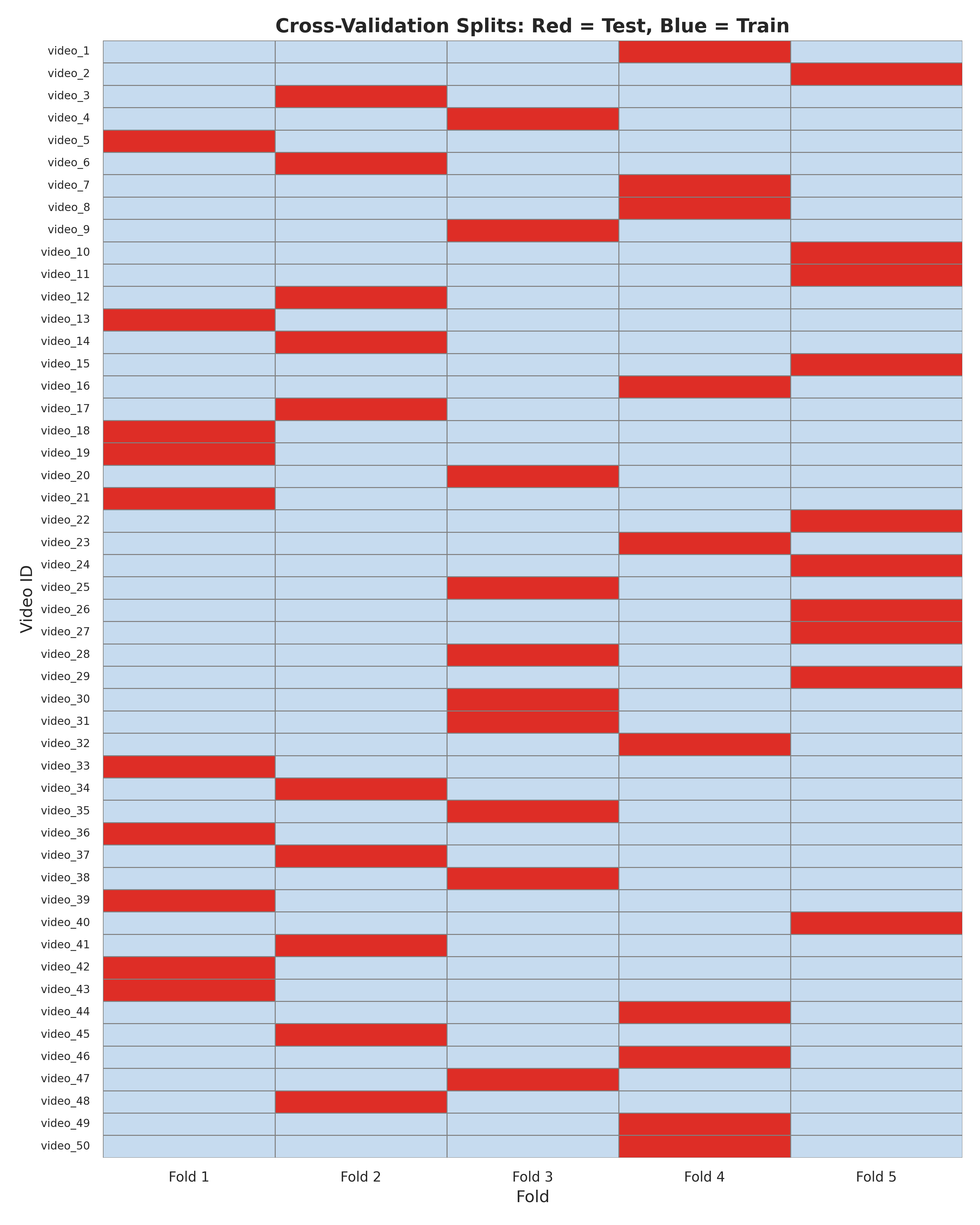}
    \caption{Cross-Validation Splits used by our proposed method on TVSUM dataset \cite{tvsum_dataset}: Red = Test, Blue = Train. Each fold's test set is indicated in red, with the training set in blue.}
    \label{fig:cv_split_heatmap}
\end{figure}

In this section, we argue that our proposed method is highly reproducible especially due to the two-stage method compared to other unsupervised approaches. We highlight the main reasoning behind that as follows,

\begin{itemize}
    \item \textbf{Network Initialization: } Our unsupervised method (stage 2) and other existing methods initialize the network either randomly or from a distribution across 5 splits on a cross-validation. However, our two-stage method has a significant advantage over all other supervised and unsupervised methods since self-supervised learning allows the same network initialization across all splits during cross validation. This allows very little standard deviation in terms of network performance not only across 5 splits but also across 10 random seeds. This has been illustrated in Fig.~\ref{fig:reproducibilty}.
    \item Additionally, we also share our non-overlapping train and test sets for videos towards highly reproducible results. Fig.~\ref{fig:cv_split_heatmap} illustrates the random splits used both in training and test sets across cross-validation and Table ~\ref{tab:video_id_filename} shows the video file names for corresponding video IDs used.
\end{itemize}

\begin{figure}[h!]
    \centering
    \includegraphics[width=0.85\textwidth]{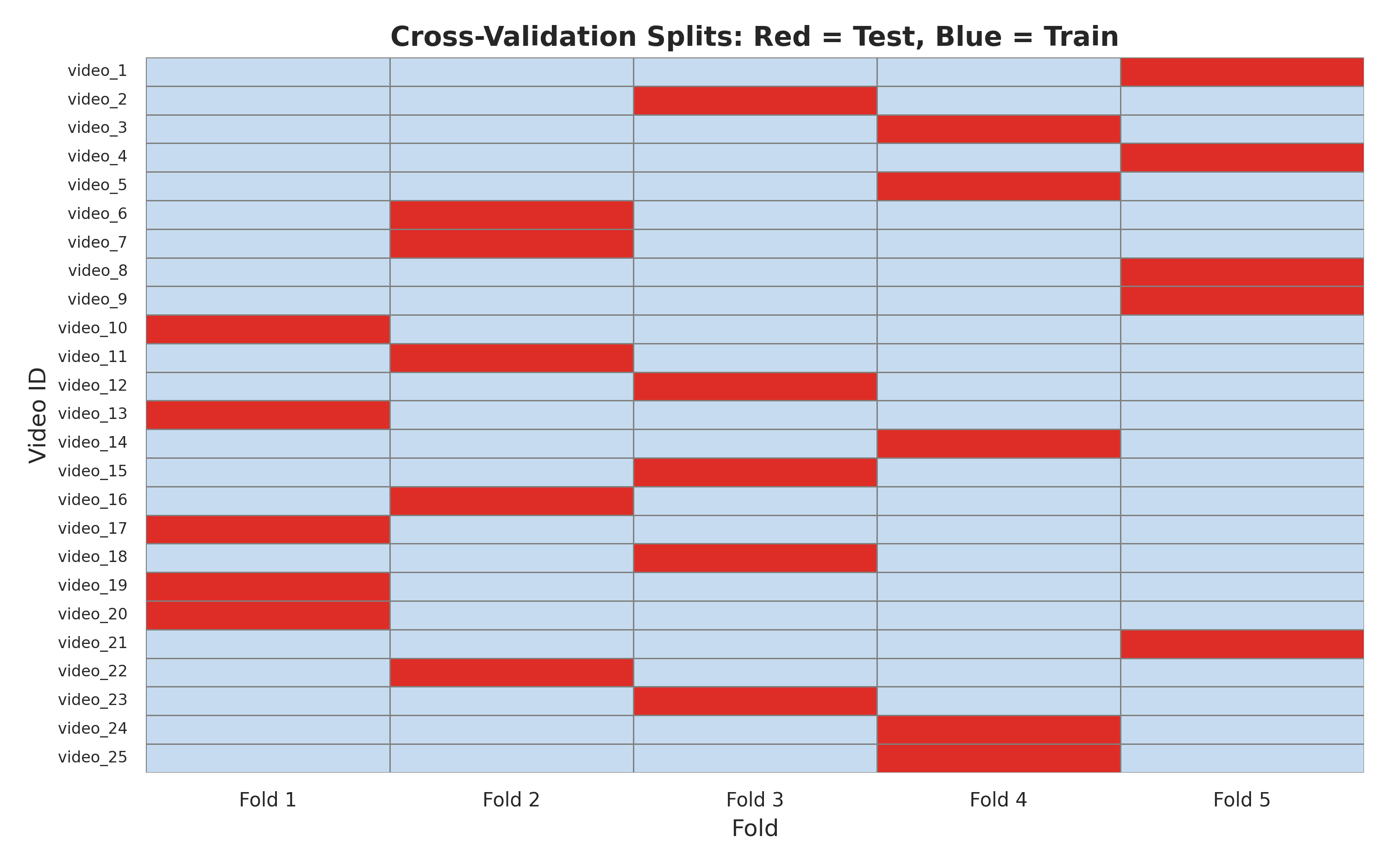}
    \caption{Cross-Validation Splits used by our proposed method on SUMME dataset \cite{summe_dataset}: Red = Test, Blue = Train. Each fold's test set is indicated in red, with the training set in blue.}
    \label{fig:cv_split_heatmap_summe}
\end{figure}


\begin{table}[h]
    \centering
    \caption{Mapping Video\_IDs to their corresponding filenames on SUMME dataset \cite{summe_dataset}.}
    \begin{tabular}{rlrl}
        \toprule
        \textbf{Video ID} & \textbf{Filename} & \textbf{Video ID} & \textbf{Filename} \\
        \midrule
            1& Air\_Force\_One&14& Notre\_Dame\\
            2& Base\_jumping&15& Paintball\\
            3& Bearpark\_climbing&16& Playing\_on\_water\_slide\\
            4& Bike\_Polo&17& Saving\_dolphines\\
            5& Bus\_in\_Rock\_Tunnel&18& Scuba\\
            6& Car\_railcrossing&19& St\_Maarten\_Landing\\
            7& Cockpit\_Landing&20& Statue\_of\_Liberty\\
            8& Cooking&21& Uncut\_Evening\_Flight\\
            9& Eiffel\_Tower&22& annotation.csv\\
            10& Excavators\_river\_crossing&23& car\_over\_camera\\
            11& Fire\_Domino&24& paluma\_jump\\
            12& Jumps&25& playing\_ball\\
            13& Kids\_playing\_in\_leaves&&\\           
        \bottomrule
    \end{tabular}
    \label{tab:video_id_filename_summe}
\end{table}


\begin{table}[h]
    \centering
    \caption{Mapping Video\_IDs to their corresponding filenames on TVSUM dataset \cite{tvsum_dataset}.}
    \begin{tabular}{rlrl}
        \toprule
        \textbf{Video ID} & \textbf{Filename} & \textbf{Video ID} & \textbf{Filename} \\
        \midrule
            1& AwmHb44\_ouw.mp4&26& 91IHQYk1IQM.mp4\\
            2& 98MoyGZKHXc.mp4&27& RBCABdttQmI.mp4\\
            3& J0nA4VgnoCo.mp4&28& z\_6gVvQb2d0.mp4\\
            4& gzDbaEs1Rlg.mp4&29& fWutDQy1nnY.mp4\\
            5& XzYM3PfTM4w.mp4&30& 4wU\_LUjG5Ic.mp4\\
            6& HT5vyqe0Xaw.mp4&31& VuWGsYPqAX8.mp4\\
            7& sTEELN-vY30.mp4&32& JKpqYvAdIsw.mp4\\
            8& vdmoEJ5YbrQ.mp4&33& xmEERLqJ2kU.mp4\\
            9& xwqBXPGE9pQ.mp4&34& byxOvuiIJV0.mp4\\
            10& akI8YFjEmUw.mp4&35& \_xMr-HKMfVA.mp4\\
            11& i3wAGJaaktw.mp4&36& WxtbjNsCQ8A.mp4\\
            12& Bhxk-O1Y7Ho.mp4&37& uGu\_10sucQo.mp4\\
            13& 0tmA\_C6XwfM.mp4&38& EE-bNr36nyA.mp4\\
            14& 3eYKfiOEJNs.mp4&39& Se3oxnaPsz0.mp4\\
            15& xxdtq8mxegs.mp4&40& oDXZc0tZe04.mp4\\
            16& WG0MBPpPC6I.mp4&41& qqR6AEXwxoQ.mp4\\
            17& Hl-\_\_g2gn\_A.mp4&42& EYqVtI9YWJA.mp4\\
            18& Yi4Ij2NM7U4.mp4&43& eQu1rNs0an0.mp4\\
            19& 37rzWOQsNIw.mp4&44& JgHubY5Vw3Y.mp4\\
            20& LRw\_obCPUt0.mp4&45& iVt07TCkFM0.mp4\\
            21& cjibtmSLxQ4.mp4&46& E11zDS9XGzg.mp4\\
            22& b626MiF1ew4.mp4&47& NyBmCxDoHJU.mp4\\
            23& XkqCExn6\_Us.mp4&48& kLxoNp-UchI.mp4\\
            24& GsAD1KT1xo8.mp4&49& jcoYJXDG9sw.mp4\\
            25& PJrm840pAUI.mp4&50& -esJrBWj2d8.mp4\\
        \bottomrule
    \end{tabular}
    \label{tab:video_id_filename}
\end{table}

\clearpage

\end{document}